%% file: main.tex
\documentclass[conference]{IEEEtran}
\IEEEoverridecommandlockouts
\usepackage{cite}
\usepackage{amsmath,amssymb,amsfonts}
\usepackage{algorithmic}
\usepackage{graphicx}
\usepackage{textcomp}
\usepackage{booktabs}
\usepackage{xcolor}
\usepackage{multirow}
\usepackage{float}
\usepackage{hyperref}
\usepackage{tikz}
\usepackage{csquotes}
\usepackage{xcolor}
\begin{document}

\title{Large Language Models Are Zero-Shot Text Classifiers}

\author{\IEEEauthorblockN{Zhiqiang Wang, Yiran Pang, Yanbin Lin}
\IEEEauthorblockA{\textit{Department of Electrical Engineering and Computer Science, Florida Atlantic University}\\
Boca Raton, FL 33431, USA \\
{\{zwang2022, ypang2022,  liny2020\}@fau.edu}
}}

\maketitle
\thispagestyle{plain}
\pagestyle{plain}

\begin{abstract}
Retrained large language models (LLMs) have become extensively used across various sub-disciplines of natural language processing (NLP). In NLP, text classification problems have garnered considerable focus, but still faced with some limitations related to expensive computational cost, time consumption, and robust performance to unseen classes. With the proposal of chain of thought prompting (CoT), LLMs can be implemented using zero-shot learning (ZSL) with the step-by-step reasoning prompts, instead of conventional question-and-answer formats. The zero-shot LLMs in the text classification problems can alleviate these limitations by directly utilizing pre-trained models to predict both seen and unseen classes. Our research primarily validates the capability of GPT models in text classification. We focus on effectively utilizing prompt strategies to various text classification scenarios. Besides, we compare the performance of zero-shot LLMs with other state-of-the-art text classification methods, including traditional machine learning methods, deep learning methods, and ZSL methods. Experimental results demonstrate that the performance of LLMs underscores their effectiveness as zero-shot text classifiers in three of the four datasets analyzed. The proficiency is especially advantageous for small businesses or teams that may not have extensive knowledge in text classification.
\end{abstract}


\section{Introduction}
%\textbf{Provide an overview of text classification, its importance, and common applications (like sentiment analysis and spam detection).Introduce LLMs, emphasizing their recent advancements and potential as zero-shot learners.State the objective of your research clearly.}

Text classification is considered the most fundamental and crucial task in the field of natural language processing. Over the past several decades, text classification issues have received extensive attention and have been effectively tackled in numerous practical applications\cite{jiang2018text,liu2020tensor,singh2022novel}. These applications contain sentiment analysis\cite{liu2022sentiment}, topic labeling \cite{chen2020dirichlet}, question answering\cite{minaee2021deep} and dialog act classification\cite{qin2020dcr}. Due to the information explosion, processing and classifying large amounts of text data manually is a time-consuming and challenging. Therefore, introducing machine learning methods for text classification tasks is indispensable.

The majority of text classification systems can typically be separated into four key stages: Feature Extraction, Dimension Reduction, Classifier Selection, and Evaluation\cite{kowsari2019text}. 
%The prevalent methods for feature extraction in text analysis include Term Frequency-Inverse Document Frequency, Term Frequency\cite{salton1988term}, Word2Vec\cite{goldberg2014word2vec}, and Global Vectors for Word Representation\cite{pennington2014glove}. Each of these techniques offers a unique way of representing text data, making them crucial tools in natural language processing and machine learning for handling and analyzing large volumes of textual information. The most frequently used methods for dimensionality reduction are Principal Component Analysis (PCA), Linear Discriminant Analysis (LDA), and Non-Negative Matrix Factorization (NMF)\cite{kowsari2019text}. Implementing dimensionality reduction techniques for pre-processing can enhance the efficiency of text classification systems.
For the classifier, traditional and popular machine learning (ML) methods include logistic regression (LR), Multinomial Naive Bayes (MNB), Logistic Regression (LG), k-nearest neighbor (KNN), Support Vector Machine (SVM), Decision Tree (DT), Random Forest, and Adaboost. These supervised ML methods face significant limitations, primarily that it requires extensive amounts of task-specific, labeled data for training before it can effectively predict outcomes on new data\cite{sarker2021machine}. Another drawback of supervised learning is that models are limited to classifying data into known classes, and are incapable of categorizing data into classes that are unseen or not labeled in the training data\cite{wang2019survey}. Recently, deep learning (DL) methods have outperformed earlier ML algorithms in natural language processing (NLP) tasks. The success of these deep learning algorithms is attributed to their ability to model intricate and non-linear relationships in data\cite{lecun2015deep}. The deep learning models for text and document classification typically involve three fundamental architectures, which are deep neural networks (DNN), recurrent neural network (RNN), and convolutional neural network (CNN). RNN usually works through Long Short-Term Memory networks (LSTM) or Gated Recurrent Units (GRU) architectures for text classification, encompassing an input layer (word embedding), hidden layers, and the output layer\cite{sutskever2011generating}. However, DL methods also need dataset labeling and a lot of training data. 
A dataset labeled with one specific set of classes cannot be repurposed to train a method designed to predict an entirely different set of classes.

Apart from these ML-based and DL-based methods, recent progress in NLP has resulted in the development of large language models (LLMs) such as Llama2, ChatGPT. Existing research\cite{chen2021evaluating,kasneci2023chatgpt,zhou2022large} suggests that the impressive performance of LLMs holds promise for their potential to address the above limitations. LLMs represent sophisticated language models characterized by their extensive parameter sizes and remarkable learning abilities\cite{chang2023survey}. The effectiveness of LLMs is commonly credited to their proficiency in few-shot or zero-shot learning within context. Pre-trained LLMs are extensively employed across various sub-disciplines of NLP and are typically recognized for their exceptional ability to learn from a few examples, which is Few-Shot Learning (FSL)\cite{wang2020generalizing}. In NLP tasks, both zero-shot learning (ZSL)\cite{xian2017zero} and FSL are techniques used to enable models to perform tasks they haven't been explicitly trained on, but they differ in their approach and reliance on training data. In FSL, the model is trained with a handful of labeled task-specific examples\cite{alhoshan2022zero}. Different from FSL, ZSL directly utilizes pre-trained models to predict both known (seen) and unknown (unseen) classes without requiring any labeled training instances and fine-tuning\cite{larochelle2008zero}. FSL is advantageous, even with access to larger datasets, because labeling data is time-consuming and training on extensive data sets can be computationally costly\cite{kadam2020review}. While ZSL can save more computational cost and time consumption with totally skipping the steps of labeling, tokenization, data pre-processing and feature extraction\cite{alhoshan2023zero}. 

ZSL is an emerging learning paradigm, aiming to tackle a task in the absence of any training examples specific to that task. The LLMs using ZSL can solve various tasks by merely conditioning the models on instructions to describe the task, which is known as “prompting”\cite{liu2023pre}. The prompts can be designed manually\cite{reynolds2021prompt} or automatically\cite{shin2020autoprompt}. GPT-3 was assessed on tasks using zero-shot, one-shot, and n-shot prompts, which included only a natural language description, one solved example, and n solved examples, respectively\cite{brown2020language,wei2021finetuned,gao2020making}. They found that GPT-3's zero-shot performance significantly lags behind its few-shot performance in tasks like reading comprehension, question answering, and natural language inference\cite{wei2021finetuned}. A potential reason for this is that without few-shot examples, it becomes more challenging for models to perform effectively on prompts that deviate from the format of the pre-training data, as same as GPT-3.5. The chain of thought prompting (CoT) was proposed in \cite{wang2022self} to feed LLMs with the step-by-step reasoning examples, instead of conventional question-and-answer formats. Zero-shot-CoT, a new approach was introduced in \cite{kojima2022large} that significantly enhanced zero-shot performance of LLMs in various reasoning tasks, such as arithmetic, symbolic reasoning, and logical reasoning, without the need for task-specific few-shot examples. By adding a simple prompt ``Let’s think step by step" before answers, the study shows that LLMs can significantly outperform standard zero-shot methods in diverse reasoning tasks. Img2LLM, a plug-and-play module was proposed by \cite{guo2023images} to generate effective LLMs prompts, describing image content as exemplar question-answer pairs. These designed prompts enabled LLMs to perform zero-shot visual question-answering without end-to-end training. The capability of LLMs was investigated in \cite{pearce2023examining} for zero-shot vulnerability repair in coding, like OpenAI’s Codex and AI21’s Jurassic J-1. A multi-turn question-answering framework for zero-shot information extraction, called as ChatIE, was introduced by \cite{wei2023zero} to leverage the capabilities of ChatGPT. 

Although there are many researches focus on the performance of zero-shot LLMs, fewer researches compare the classification performance of zero-shot LLMs with traditional ML methods, DL methods, and ZSL methods. For an intuitive comparison, we design some step-by-step prompts to analyze the zero-shot classification performance of state-of-the-art large language models, like Llama2, GPT-3.5, and GPT-4. We carry out comprehensive assessments of these models using four different datasets, including the applications of the sentiment analysis, a four-class classification task, and the spam detection. Various ML methods, DL methods, and ZSL methods are implemented to compare performance, such as MNB, LG, RF, DT, and KNN, RNN, LSTM, GRU, BART\cite{lewis2019bart} and DeBERTa\cite{he2020deberta}. Our main contribution is to evaluate the zero-shot performance of LLMs against existing other text classification approaches.

Our main contributions are as follows:
\begin{itemize}
    \item \textbf{Innovative Application of GPT Models in Text Classification}: We demonstrate how GPT models can simplify the text classification process by directly generating classification labels, thereby avoiding traditional feature extraction and classifier training steps.
    \item \textbf{Extensive Evaluation and Comparison Across Multiple Datasets}: Our research includes a wide-ranging evaluation across various domain datasets, comparing the performance of GPT models with traditional machine learning methods and neural network models, affirming their effectiveness in text classification tasks.
    \item \textbf{Practical Implications for Small Businesses or Teams and Open Source Contribution}: We highlight the practical value of GPT models in text classification for small businesses or teams, who may lack in-depth knowledge in this area. To foster further research and application in this field, we have made our code open source, allowing community members to directly utilize and improve upon these models.
\end{itemize}

The organization of the paper is summarized as follows: In Section II, we briefly discuss the background and related work. Section III presents our proposed methodology, mainly including the overview of proposed method and practical methodology. Section IV presents the experiment results of all methods using four different datasets. Section V discusses the results. Section VI concludes our paper. Finally, Section VII states the data and results availability.

\section{Background and Related Work}
% Discuss traditional text classification methods (NB, SVM, RNN, GRU, LSTM, etc) and their common challenges.
% They may perform well in one use case when it applys to another field, a new model need to be trained
% LLM does not require that, it is able to take over different text classification tasks at the same time
% Explore the evolution of LLMs, focusing on GPT-3.5 Turbo and GPT-4, and their capabilities in text classification without explicit training.
% Review existing literature on zero-shot learning and its applications in NLP.
\input{background}

\input{method}

\begin{figure*}[ht]
  \centering
  \includegraphics[width=0.9\textwidth]{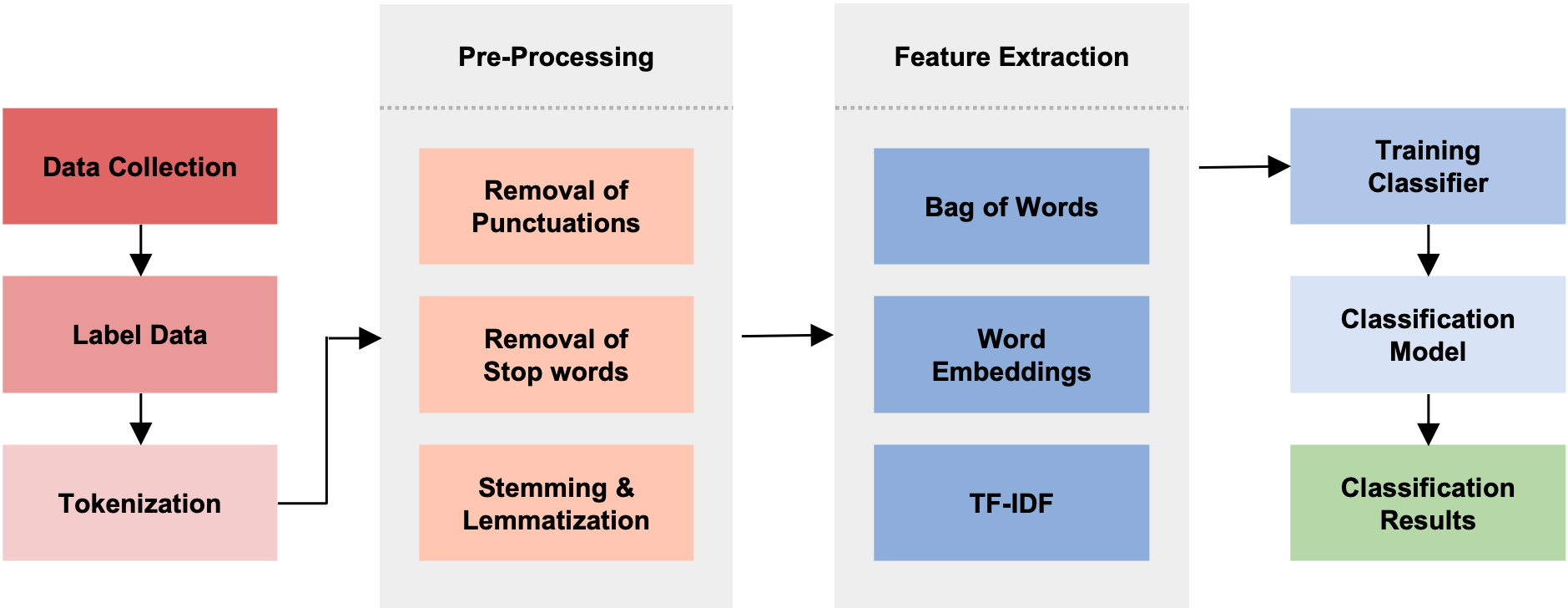}
    \caption{Traditional text classification flow}
    \label{fig:traditional_flow}
\end{figure*}

\begin{figure*}[ht]
  \centering
  \includegraphics[width=0.6\textwidth]{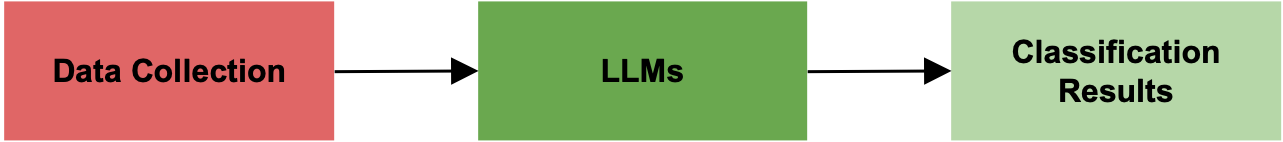}
    \caption{LLMs' zero shot text classification flow}
    \label{fig:llms_flow}
\end{figure*}

\begin{table*}[ht]
\centering
\caption{Examples of Covid19 Tweets with Different Sentiment Labels}
\renewcommand{\arraystretch}{1.3}
\begin{tabular}{c|p{15cm}}
\hline
\textbf{Label} & \textbf{Tweet} \\
\hline
Neural & Vistamalls says supermarket sales to 'balance' \#COVID19 impact \url{https://t.co/caE2rT6MhO} \\
\hline
Negative & Just now on the telly, Woolies have stopped all online and click n collect orders. Due to overwhelming demand. \#coronavirus \#StopPanicBuying \\
\hline
Positive & Efforts 2 contain \#COVID-19 are shifting demand \& disrupting Ag supply chains. @raboresearch has collated our analysis of current \& expected impacts in one place 2 help our @RabobankAU network keep informed. \url{https://t.co/P41vjG4uD6} \\
\hline
\end{tabular}
\label{tab:tweets}
\end{table*}

\begin{table*}[ht]
\centering
\caption{Examples of e comerce text with Different Labels}
\renewcommand{\arraystretch}{1.3}
\begin{tabular}{c|p{14cm}}
\hline
\textbf{Label} & \textbf{E commerce Text} \\
\hline
Clothing \& Accessories & Cherokee by Unlimited Boys' Straight Regular Fit Trousers Cherokee kids beige trousers made of 100\% cotton twill fabric. \\
\hline
Household & Nutella Hazelnut Spread with Cocoa, 290g Size:290g   Because the taste is simply unique! The secret is its special recipe, the selected ingredients and the careful preparation. Here we want to tell you about Nutella and all the passion and care that we put in its production every day.\\
\hline
Books & NIACL Assistant Preliminary Online Exam Practice Work Book - 2280.  \\
\hline
Electronics & Transcend 512 MB Compact Flash (TS512MCF300) Transcend's CF300 cards are high-speed industrial CF cards offering impressive 300X transfer rates. With matchless performance and durability, CF300 CF cards are perfect for POS and embedded systems that require both industrial-grade reliability and an ultra-high speed data transfer.\\
\hline
\end{tabular}
\label{tab:comerce}
\end{table*}

\section{Experiment and Results}
While accuracy calculation is the most straightforward evaluation method, it is not effective for unbalanced datasets\cite{huang2005using}. F1 Score, Matthews Correlation
Coefficient (MCC), Accuracy (ACC), receiver operating characteristics (ROC), and area under the ROC curve (AUC) methods are suitable for text classification algorithms' evaluation\cite{kowsari2019text}.

In this study, we conduct an extensive evaluation of our proposed methodologies across four distinct datasets. These datasets encompass a diverse range of applications: sentiment analysis was performed using COVID-19 related tweets (Gabriel Preda, 2020)\cite{gabrielpreda_2020} and economic texts (Malo et al., 2014)\cite{malo2014good}, a four-class classification task was applied to e-commerce texts (Gautam, 2019)\cite{gautam_2019_335582}, and spam detection was implemented on an SMS dataset (SMS Spam Collection, 2012)\cite{misc_sms_spam_collection_228}.

For each dataset, a comprehensive set of models is employed to assess the effectiveness of the proposed methods. These models are spanned to three categories:

\begin{itemize}
    \item \textbf{Traditional ML Algorithms:} This category includes MNB, LG, RF, DT, and KNN.
    \item \textbf{DL Architectures:} We utilize advanced deep neural network models such as RNN, LSTM, and GRU.
    \item \textbf{ZSL Models:} In this category, we explore the performance of zero-shot models, specifically the transformer-based models, BART (facebook/bart-large-mnli) and DeBERTa (microsoft/deberta-large-mnli). 
    \item \textbf{LLMs:} State-of-the-art large language models including \textbf{Llama2 }(Llama2-70B), \textbf{GPT-3.5 }(gpt-3.5-turbo-1106), and \textbf{GPT-4 }(gpt-4-1106-preview) were assessed.
\end{itemize}

For all traditional ML algorithms and DL models, we maintain uniformity in the input processing. This means that each model uses the same processed text derived from a consistent raw text processing flow, encompassing both training and testing datasets. This approach ensures that variations in performance could be attributed more directly to the model’s capabilities rather than differences in input processing.

On the other hand, for zero-shot learning models and LLMs, we directly employ the raw text from the testing dataset. It's important to note that the testing dataset remained identical across all models, fostering a fair and consistent basis for comparison.

It is worth mentioning that the traditional ML algorithms and DL models can not undergo any specialized or model-specific text processing enhancements. This decision is intentional to minimize the number of variables influencing the experimental results. While this might have resulted in performance that is not state-of-the-art for each individual model, it is crucial for maintaining the integrity of the comparative analysis. Our goal is to evaluate each model's inherent capabilities under standardized conditions, thereby providing a more transparent and direct comparison of their performance in the context of zero-shot text classification.

In order to maintain consistency and precision in the results obtained from LLMs, we standardize the hyper-parameters across all LLMs by setting the temperature to 0.01 and the top$_p$ to 0.9. The temperature parameter controls the randomness in the prediction distribution, with a lower temperature resulting in less random completions. The top$_p$ parameter, also known as nucleus sampling, restricts the model's choices to the top 90\% probabilities, thereby preventing the selection of highly improbable words. At the same time, the prompt used for different LLMs for the same dataset is also the same. 

Furthermore, to ensure uniformity in model inputs, the prompts for the same dataset are kept identical across different LLMs. For different datasets, the same core structure of the prompts are used that only labels and dataset names are adjusted as required. This approach is adopted to minimize variability in model responses attributable to differences in input. Such an approach enables a more precise evaluation of each model's performance to focus on their inherent capabilities rather than variances in input.

\begin{table}[h!]
\centering
\caption{\small{Results in Sentiment Classification}}
\renewcommand{\arraystretch}{1.3}
\begin{tabular}{lccc|ccc}
\toprule
& \multicolumn{3}{c}{COVID19 Tweet} & \multicolumn{3}{c}{Economic Text} \\
\cmidrule(r){2-4} \cmidrule(r){5-7}
& ACC & F1 & AUC & ACC & F1 & AUC \\
\midrule
MNB          & 0.3933 & 0.3639 & 0.5531 & 0.4533 & 0.3632 & 0.5563 \\
LG           & 0.4333 & 0.3488 & 0.5404 & 0.5200 & 0.3066 & 0.5427 \\
RF           & 0.4467 & 0.3184 & 0.6184 & 0.5133 & 0.3453 & 0.5990 \\
DT          & 0.4733 & 0.4105 & 0.5602 & 0.4067 & 0.3446 & 0.5060 \\
KNN          & 0.3800 & 0.3486 & 0.5216 & 0.4800 & 0.3620 & 0.5614 \\
\hline
RNN          & 0.7400 & 0.7186 & 0.8925 & 0.6333 & 0.5797 & 0.7874 \\
LSTM         & 0.7867 & 0.7619 & 0.8925 & 0.6533 & 0.4627 & 0.7293 \\
GRU          & \textbf{0.8200} & \textbf{0.8106} & 0.9226 & 0.6933 & 0.5767 & 0.7928 \\
\hline
BART         & 0.5000 & 0.3516 & 0.5882 & 0.4600 & 0.4258 & 0.6603 \\
DeBERTa      & 0.5467 & 0.3805 & 0.5954 & 0.4467 & 0.4251 & 0.6385 \\
\hline
Llama2       & 0.5267 & 0.4748 & -      & 0.7000 & 0.5230 & - \\
GPT-3.5      & 0.5333 & 0.4943 & -      & 0.6667 & 0.6683 & - \\
GPT-4        & 0.5267 & 0.5095 & -      & \textbf{0.7133} & \textbf{0.7096} & - \\
\bottomrule
\end{tabular}
\label{tab:sentiment}
\end{table}

From Table \ref{tab:sentiment}, it shows that the performance of the evaluated algorithms and models in sentiment classification tasks for both COVID-19 tweets and economic texts is not outstanding. Within the four categories of classifiers, traditional ML algorithms consistently present the least favorable accuracy across the two datasets.

Notably, the accuracies of most models do not surpass the 80\% threshold, with the exception of the GRU model, which achieves the highest accuracy of 82\% in classifying COVID-19 tweets. This represents a substantial increase of approximately 20-30\% over the zero-shot classifiers and LLMs, which approach around 50-55\%. Furthermore, these results highlight a considerable improvement of about 3-10\% over the traditional ML algorithms.

In the analysis of economic texts, while DL methods perform better than traditional ML algorithms, the margin is narrower compared to the results with COVID-19 tweets. Particularly that among the 3 LLMs, Llama2 and GPT-4 slightly outperform all the other algorithms or models with the accuracies of 70.00\% and 71.33\% respectively that GPT-4 achieves the highest accuracy.

\begin{figure*}[ht]
  \centering
  \includegraphics[width=0.95\textwidth]{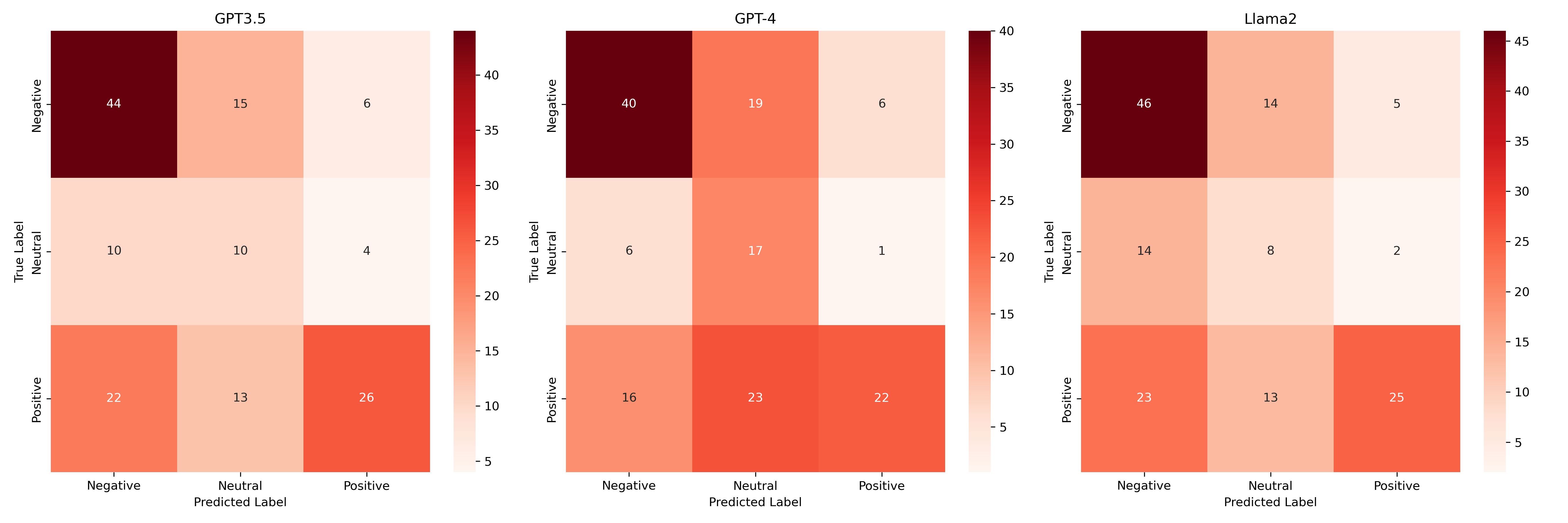}
    \caption{The confusion matrices for LLMs' classification results in COVID19 tweets.}
    \label{fig:tweets_matrix}
\end{figure*}

\begin{table}[h!]
\centering
\caption{\small{Results in original tweets V.S. Clean tweets }}
\renewcommand{\arraystretch}{1.3}
\begin{tabular}{lc|c|c}
\toprule
& \multicolumn{1}{c}{GPT-3.5} & \multicolumn{1}{c}{GPT-4} & \multicolumn{1}{c}{LlaMa2}\\
\hline
Original Text    & \textbf{0.5413$\pm$0.0099} & 0.5200$\pm$0.0047 & 0.5280$\pm$0.0099  \\
Clean Text       & 0.5145$\pm$0.0087 & \textbf{0.5560$\pm$0.0203} & 0.4973$\pm$0.0037 \\
\bottomrule
\end{tabular}
\label{tab:clean_tweets}
\end{table}

In COVID Tweet testing dataset, there are 65 negative, 24 neutral, and 61 positive tweets. By looking into the details of LLMs predicted results, GPT-3.5 and Llama2 models show a preference towards predicting tweets as "negative", more than what the original distribution suggests. While GPT-4 tends to classifier tweets into "neutral". 

Considering tweets usually includes, urls, html tags, hash tags mentions and text pre-handling is an important step in traditional text classification process, we remove urls, htmls tags, digits, hash tags, mentions and stop words from tweets dataset. Then the cleansed tweets are re-evaluated in LLMs over 5 times to ensure the results are reliable and consistent.

Table \ref{tab:clean_tweets} shows that before text cleaning, GPT-3.5 has superior accuracy, while for post-cleaning, GPT-4 outperforms the others. The decreased performance in GPT-3.5 and LlaMa2 after cleaning suggests that these models may leverage the full range of information contained in raw tweets, including hashtags and mentions, to inform their predictions. Conversely, GPT-4's improved accuracy indicates a possible advantage in processing cleaner, more structured data. This variation in model performance addresses the importance of considering the specific characteristics of text data and the corresponding pre-processing steps when working with different LLMs.

\begin{table}[h!]
\centering
\caption{\small{Results in E Commerce Text}}
\renewcommand{\arraystretch}{1.3}
\begin{tabular}{lccc}
\toprule
& \multicolumn{3}{c}{E commerce text} \\
\cmidrule(r){2-4}
& ACC & F1 & AUC \\
\midrule
MNB          & 0.2667 & 0.2546 & 0.5289 \\
LG           & 0.3867 & 0.2820 & 0.6529 \\
RF           & 0.5133 & 0.4410 & 0.7267 \\
DT          & 0.5467 & 0.5412 & 0.6964 \\
KNN          & 0.3600 & 0.3159 & 0.5891 \\
\hline
RNN          & \textbf{0.9600} & \textbf{0.9598} & 0.9964 \\
LSTM         & 0.9467 & 0.9468 & 0.9854 \\
GRU          & 0.9400 & 0.9401 & 0.9883 \\
\hline
BART         & 0.7133 & 0.7272 & 0.4391 \\
DeBERTa      & 0.6267 & 0.6358 & 0.4726 \\
\hline
Llama2       & 0.8067 & 0.6644 & - \\
GPT-3.5      & 0.8867 & 0.8935 & - \\
GPT-4        & 0.9000 & 0.9078 & - \\
\bottomrule
\end{tabular}
\label{tab:e_commerce}
\end{table}

\begin{table}[h!]
\centering
\caption{\small{Results in SMS}}
\renewcommand{\arraystretch}{1.3}
\begin{tabular}{lccc}
\toprule
& \multicolumn{3}{c}{SMS} \\
\cmidrule(r){2-4}
& ACC & F1 & AUC\\
\midrule
MNB          & 0.7600 & 0.6425 & 0.7346 \\
LG           & 0.8333 & 0.4545 & 0.4808 \\
RF           & 0.9067 & 0.7678 & 0.7346 \\
DT          & 0.8667 & 0.7337 & 0.7538 \\
KNN          & 0.8400 & 0.6384 & 0.6327 \\
\hline
RNN          & 0.9800 & 0.9558 & 0.9462 \\
LSTM         & 0.9600 & 0.9005 & 0.8500 \\
GRU          & \textbf{0.9867} & \textbf{0.9699} & 0.9500 \\
\hline
BART         & 0.7000 & 0.5479 & 0.5942 \\
DeBERTa      & 0.8200 & 0.6906 & 0.7481 \\
\hline
Llama2       & 0.7267 & 0.4441 & - \\
GPT-3.5      & 0.8733 & 0.7996 & - \\
GPT-4        & 0.9733 & 0.9467 & - \\
\bottomrule
\end{tabular}
\label{tab:sms}
\end{table}

In table \ref{tab:e_commerce}, The DL models demonstrate superior performance with all the accuracies surpassing 90\% where RNN achieves the highest accuracy, 0.9600.  This suggests that RNN architectures are particularly effective for this task. Notably, the LLMs also perform well with GPT-4 achieving the highest accuracy of 0.9000, which addresses the effectiveness of LLMs in understanding and classifying complex e-commerce text.

In table \ref{tab:sms}, the DL models again show strong results in detecting spam SMS, particularly in accuracy in F1 score with all accuracies over 95\% which are even better in classifying e commerce text. Traditional ML algorithms, like RF and DT also show commendable performance, especially in accuracy. As for LLMs, GPT-4 showcases a significant lead in accuracy at 97.33\% which is only slightly worse than GRU, 98.67\% and RNN, 98.00\% but better than LSTM, 96.00\%.

\section{Discussion}

In this study, GPT-4 consistently outperformes traditional ML algorithms across all four datasets. It is also worth pointing out that Llama2 and GPT-3.5 show strengths in sentiment analysis and e-commerce text classification. More importantly, Llama2 and GPT-4 defect all the other models in economic text analysis. Despite the gap in accuracy between the LLMs and DL models in COVID-19 sentiment analysis, LLMs delivers robust results in the remaining tasks. 

Base on our research and experiment, it shows that while setting a low temperature and a high top$_p$ value, LLMs might not always yield the expected output. For instance, despite given prompts, Llama2 occasionally adds extraneous text or generated more outputs than inputs. In contrast, traditional ML and DL models typically produce standardized outputs, facilitating downstream tasks. The latest OpenAI LLMs offer JSON output format support and consistent result through the use of a constant random seed that we hope to see in other LLMs shortly. 

However, a concern arises from the LLMs' training dataset, which usually includes most of the open-source internet data, that it may have included our evaluation datasets, potentially biasing the results. Nonetheless, in our experience with commercial private data, GPT-3.5 and GPT-4 have demonstrated commendable performance, comparable or superior to RNNs/CNNs, with accuracies exceeding 90\%.

\section{Conclusion and Future Work}

The performance of LLMs in three out of the four datasets studies supports the conclusion that LLMs can effectively function as zero-shot text classifiers. This capability is particularly beneficial for small businesses or teams lacking in-depth expertise in text classification. It enables them to rapidly deploy text classifiers, allowing them to concentrate on downstream tasks. 

Future accuracy improvements might include refining prompts with more detailed background information or more precise label definitions. Another prospective improvement could involve implementing a critic agent, drawing inspiration from actor-critic algorithms\cite{konda1999actor}, to evaluate and enhance the results provided by LLMs. This study opens up new approaches, especially in sentiment analysis where none of the algorithms achieved superior accuracy with a standard text classification process, indicating a promising direction for future research.

\section{Data and Results Availability Statement}
Source code, datasets and all experiment logs generated and/or analysed in this study are available in the following GitHub repository: https://github.com/yeyimilk/llm-zero-shot-classifiers.

\bibliographystyle{IEEEtran}
\bibliography{ref}

\end{document}

%% file: background.tex
\subsection{Rule-Based Methods}
Early work in text classification primarily focused on rule-based methods. Decision tree algorithms, including C4.5\cite{quinlan2014c4} and CART\cite{loh2011classification}, classify texts by constructing tree structures based on feature selection criteria. These methods are easy to understand and implement. However, they may generate overly complex structures when dealing with complex or high-dimensional data, which leads to overfitting issues. Expert systems, like MYCIN\cite{shortliffe2012computer}, make decisions based on rules defined by domain experts. Pattern matching methods identify specific types of texts by matching predefined patterns or keyword sequences. They perform well in applications such as spam email filtering\cite{androutsopoulos2000learning}. However, these rule-based approaches depend heavily on initial settings and may struggle to adapt to new patterns or noisy data.
\subsection{Probability-based Methods}
Compared to rule-based models, probability-based models offer greater flexibility and generalization capabilities through their mathematical frameworks and data-driven approach. The Naive Bayes classifier\cite{xu2018bayesian}, employing Bayes' theorem for text classification, stands out even with its assumption of feature independence. It demonstrates significant effectiveness in areas like spam detection and sentiment analysis. The Naive Bayes classifier is widely acknowledged for its simplicity and robust performance with large datasets. The Hidden Markov Model (HMM)\cite{rabiner1989tutorial} is another key probability-based model, especially suitable for processing sequential data. In natural language processing tasks such as part-of-speech tagging and speech recognition, HMMs effectively address challenges by considering state transition probabilities.
\subsection{Geometry-based Methods}
Geometry-based methods offer a distinct perspective in handling high-dimensional data. In text classification, these methods primarily focus on the spatial relationships between data points. SVM\cite{joachims2002learning} effectively classify text by finding the optimal separating hyperplane in a high-dimensional space. This approach is particularly suited for scenarios with large and complex feature spaces, as it maximizes the margin between classes to enhance classification accuracy. However, SVMs may face challenges in computational efficiency and resource consumption when dealing with very large datasets. Techniques like Principal Component Analysis (PCA)\cite{wold1987principal} and Linear Discriminant Analysis (LDA)\cite{torkkola2001linear} simplify the classification task by reducing data dimensions. These methods effectively lower complexity and computational costs but may lose important information for classification during the dimensionality reduction process.
\subsection{Statistic-based Methods}
Statistic-based models in text classification utilize the statistical properties of data for decision-making. The KNN\cite{guo2003knn} classifies a data point by analyzing its closest neighbors. However, KNN may encounter efficiency issues with large datasets as it requires calculating the distance between each data point and every other point. Logistic Regression\cite{genkin2007large} classifies by estimating the probability of data belonging to a specific category. It performs well in text classification tasks with relatively simple feature relationships. However, statistic-based models are generally sensitive to data preprocessing and feature selection. They struggle with complex or heavily nonlinear feature-rich data.
\subsection{Deep learning Methods}
Deep learning has become a key technology in text classification, capable of handling complex language features. CNN text classification model\cite{kim2014convolutional} captures local textual features through convolutional layers. This approach excels in sentiment analysis and topic categorization. LSTM\cite{sutskever2014sequence} and GRU\cite{chung2014empirical}, as optimized versions of RNNs, are particularly effective in addressing long-distance dependencies in text. Transformer models, like BERT\cite{devlin2018bert}, achieve remarkable results in various NLP tasks by utilizing self-attention mechanisms. Specifically, the BERT model demonstrates powerful capabilities in text classification tasks. However, these methods typically require substantial data for training, often necessitating extensive datasets to achieve optimal performance. This reliance on large training sets can pose challenges, especially when collecting or labeling data is difficult or impractical. 
\subsection{Zero-shot Methods}
ZSL aims to classify data without direct examples of certain classes during training. This offers a solution to the limitations of data-intensive deep learning models in text classification. Methods based on knowledge graphs, as explored in recent studies, utilize auxiliary information about inter-class relationships, represented in rich semantic knowledge graphs. These methods have been instrumental in achieving state-of-the-art performance across several benchmarks and tasks \cite{chen2021low}. Another significant advancement in ZSL is the use of semantic embedding vectors. This approach, which conceptualizes zero-shot learning as a regression problem from input to embedding space, has shown effectiveness in tasks like ImageNet zero-shot learning \cite{norouzi2013zero}. Furthermore, the exploration of large pre-trained language models (PLMs) such as BERT in zero-shot learning scenarios has opened new avenues. Research has revealed that strategies like Multi-Null Prompting in BERT family models can yield promising results, surpassing manually created prompts, although some limitations exist, particularly in language understanding tasks under zero-shot settings \cite{wang2023bert}. In addition to these methods, the adaptation of sophisticated PLMs like BART \cite{lewis2019bart} and DeBERTa \cite{he2020deberta} has further expanded the capabilities of zero-shot learning in text classification. These models, with their advanced architectures and pre-training methodologies, have been leveraged to enhance performance in various NLP tasks, including those that require understanding and generating nuanced human language. BART, with its unique combination of bidirectional and autoregressive transformers, and DeBERTa, with its disentangled attention mechanism, exemplify the advancements in leveraging deep learning for effective zero-shot text classification.

% \subsection{Zero-shot Methods}
% Zero-shot learning (ZSL) aims to classify data without direct examples of certain classes during training. This offers a solution to the limitations of data-intensive deep learning models in text classification. Methods based on knowledge graphs, as explored in recent studies, utilize auxiliary information about inter-class relationships, represented in rich semantic knowledge graphs. These methods have been instrumental in achieving state-of-the-art performance across several benchmarks and tasks \cite{chen2021low}. Another significant advancement in ZSL is the use of semantic embedding vectors. This approach, which conceptualizes zero-shot learning as a regression problem from input to embedding space, has shown effectiveness in tasks like ImageNet zero-shot learning \cite{norouzi2013zero}. Furthermore, the exploration of large pre-trained language models (PLMs) such as BERT in zero-shot learning scenarios has opened new avenues. Research has revealed that strategies like Multi-Null Prompting in BERT family models can yield promising results, surpassing manually created prompts, although some limitations exist, particularly in language understanding tasks under zero-shot settings \cite{wang2023bert}. 

%% file: method.tex
\section{Methodology}
As shown in Fig.\ref{fig:traditional_flow}, traditional text classification involves three key steps: data preprocessing, feature extraction, and classifier training. Each step plays a vital role in the overall process. Given a textual input \(x = (x_1, x_2, \ldots)\), the first step is standardizing and preprocessing. This step includes noise removal (e.g., punctuation and special characters), stop word filtering, stemming, and lemmatization. The aim is to reduce noise and standardize text data for subsequent feature extraction. The preprocessing function \(P\) can be expressed as:
\begin{equation}
    P(x) \rightarrow x'
\end{equation}
where \(x'\) denotes the preprocessed text. Next, the processed text \(x'\) undergoes feature extraction. Common techniques include bag-of-words, TF-IDF, and word embeddings. This step transforms text into a numerical feature vector suitable for machine learning models. The feature extraction function \(F\) can be defined as:
\begin{equation}
    F(x') \rightarrow h
\end{equation}
where \(h\) represents the feature vector of the text. Finally, a machine learning algorithm (e.g., Support Vector Machine, Decision Tree, Neural Network) is employed to construct a classification model. This model predicts the category of text based on the feature vector \(h\). The classification model \(C\) is formulated as:
\begin{equation}
    p(y|x') = \text{softmax}(W \cdot \text{MLP}(h))
\end{equation}
where \(W\) denotes the trainable parameters of the classifier, typically trained from scratch. In this traditional approach, each step is essential, forming a complete text classification process. While this method can achieve high accuracy, especially when fine-tuned for specific tasks, it may have limitations in processing efficiency for large datasets and adaptability to novel task types.

Utilizing GPT models for text classification employs a single-step, prompt-based method. This streamlined approach leverages GPT's generative capabilities to directly produce specific classification labels. The prompt is meticulously designed to guide the GPT model towards generating a precise classification label in a predefined format. Considering examples from Table~\ref{tab:tweets} and Table~\ref{tab:comerce}, a prompt for sentiment analysis could be structured as follows:

\begin{quote}
    \texttt{You are an AI assistant and you are very good at doing e-commerce products classification. You are going to help a customer to classify the products in the e-commerce website. You are only allowed to choose one of the following 4 categories: Household, Books, Clothing \& Accessories, Electronics. Please provide only one category for each product in JSON format where the key is the index for each product and the value is one of the 4 categories. For example: \{1: Household\}. Please do not repeat or return the content back again, just provide the category in the defined format.}
\end{quote}

This prompt explicitly instructs the GPT model to classify products into one of four categories and express the outcome in JSON format. The classification process using this prompt can be formalized as:
\begin{equation}
    \text{GPT-Response}(\text{Prompt}) \rightarrow \text{JSON Classification}
\end{equation}
Here, \(\text{GPT-Response}\) represents the GPT model processing the prompt, and \(\text{JSON Classification}\) is the output in JSON format, indicating the category for each product.

This method efficiently utilizes the natural language processing and generative capabilities of GPT models. By directing the model to produce classification results in a specific format, it simplifies the classification process and eliminates the need for intermediate steps like feature extraction or explicit verbalizer mapping. This makes it a highly practical approach for diverse text classification tasks.

% Using GPT models for text classification involves a streamlined, single-step prompt-based approach. This method harnesses the generative capabilities of GPT to directly produce classification labels such as “positive” or “negative”. In this approach, the prompt is carefully designed to elicit a direct response from the GPT model. The prompt integrates the input text with a query structure that guides the model to generate a specific classification label. Considering examples from Table~\ref{tab:tweets} and Table~\ref{tab:comerce}, a prompt for sentiment analysis could be structured as \texttt{[X] is the sentiment of this sentence positive or negative?} This direct query prompts the GPT model to respond with either “positive” or “negative”. The process can be represented as: \texttt{GPT-Response("[X] Is the sentiment of this sentence positive or negative?") $\rightarrow$ Label.} In this representation, \texttt{GPT-Response} is the function encapsulating the GPT model's processing, and \texttt{Label} is the directly generated classification outcome.

% This single-step method is highly efficient, which makes the sophisticated natural language understanding and generative capabilities of GPT models. It simplifies the classification process by eliminating intermediate steps like feature extraction or explicit verbalizer mapping, making it a practical approach for various text classification tasks.